\documentclass[letterpaper]{article} 

\usepackage{todonotes} 
\usepackage[inline]{enumitem} 

\usepackage{aaai23}  
\usepackage{times}  
\usepackage{helvet}  
\usepackage{courier}  
\usepackage[hyphens]{url}  
\usepackage{graphicx} 
\usepackage{natbib}  
\usepackage{caption} 
\usepackage{algorithm}
\usepackage{algorithmic}

\usepackage{newfloat}
\usepackage{listings}
\DeclareCaptionStyle{ruled}{labelfont=normalfont,labelsep=colon,strut=off} 
\floatstyle{ruled}
\newfloat{listing}{tb}{lst}{}
\floatname{listing}{Listing}
\title{Position Paper on Dataset Engineering to Accelerate Science}

\author {
    Emilio Vital Brazil,
    Eduardo Soares,
    Lucas Villa Real,
    Leonardo Azevedo,\\
    Vinicius Segura,
    Luiz Zerkowski, and
    Renato Cerqueira
}
\affiliations {
    IBM Research\\
    Rio de Janeiro, RJ, Brazil \\
    evital@br.ibm.com, eduardo.soares@ibm.com, lucasvr@br.ibm.com, lga@br.ibm.com,\\ vboas@br.ibm.com, luisvz@ibm.com, rcerq@br.ibm.com 
}

\usepackage{bibentry}

\begin{document}

\maketitle

\newcommand{\etal}{\textit{et al.}}
\newcommand{\ie}{\textit{i.e., }}

\begin{abstract}
Data is a critical element in any discovery process. 
In the last decades, we observed exponential growth in the volume of available data and the technology to manipulate it. 
However, data is only practical when one can structure it for a well-defined task. 
For instance, we need a corpus of text broken into sentences to train a natural language machine-learning model.
In this work, we will use the token \textit{dataset} to designate a structured set of data built to perform a well-defined task.
Moreover, the dataset will be used in most cases as a blueprint of an entity that at any moment can be stored as a table.
Specifically, in science, each area has unique forms to organize, gather and handle its datasets.
We believe that datasets must be a first-class entity in any knowledge-intensive process, and all workflows should have exceptional attention to datasets' lifecycle, from their gathering to uses and evolution.
We advocate that science and engineering discovery processes are extreme instances of the need for such organization on datasets, claiming for new approaches and tooling.
Furthermore, these requirements are more evident when the discovery workflow uses artificial intelligence methods to empower the subject-matter expert.
In this work, we discuss an approach to bringing datasets as a critical entity in the discovery process in science.
We illustrate some concepts using material discovery as a use case. 
We chose this domain because it leverages many significant problems that can be generalized to other science fields.
\end{abstract}

\section{Introduction}
Data is a critical element in any discovery process -- it appears at the beginning of the process as input for experimentation, and at the end, as evidence to support the results.
In the last few decades, we observed exponential growth in the volume of available data and the technology to generate and manipulate it (according to IDC, about 64 zettabytes were created or copied in 2020~\cite{overberg2021}).
However, data is only practical when one can use it for a well-defined task.

Besides the sheer volume, data can be unstructured and get more complex according to the domain, particularly in science applications. 
Moreover, organizing data and taking care of their entire lifecycle is especially vital when Artificial Intelligence (AI) techniques start to be critical in scientific processes.
For instance, recently, the chemical industry has augmented traditional human-intensive work with automated, parallel, and iterative processes driven by AI to accelerate the materials-discovery~\cite{Pyzer-knapp2022}.
This incorporation of AI in the materials-discovery workflow brought a set of novel problems in handling data, for example, how to qualify and filter thousand of molecule candidates created by machine learning generative techniques~\cite{Hoffman2021, tadesse2022mpego}.

Nowadays scenario, where there is a high demand to accelerate scientific discoveries, which depends on a massive quantity of data, we advocate that it is paramount to look at data from a new perspective, bringing together data and tasks to remodel the concept of the dataset.
To be used in any discovery process, data must have a well-defined structure, associations with domain knowledge, and a set of operations and analytics to evolve it. 
However, many methods to give such effects to data will strongly depend on the task we will perform with the data.
For instance, representing a molecule is an enormous issue in chemical applications, and its choice must consider the final task~\cite{OBoyle2011}.
Another common problem for working with data is how to define its lifecycle, and again we notice good practices for using the task to define it.
We must create a dataset guided by the task that we will perform.
Furthermore, while the operations executed on it are still creating information related to the original task, we consider it a natural evolution of the dataset, keeping its versioning.
In summary, we propose working with datasets as first-class entities in the discovery workflow, using the task as its main characteristic to guide its complete lifecycle.


\subsection{Our View}

\begin{figure}
    \centering
    \includegraphics[width=0.95\linewidth]{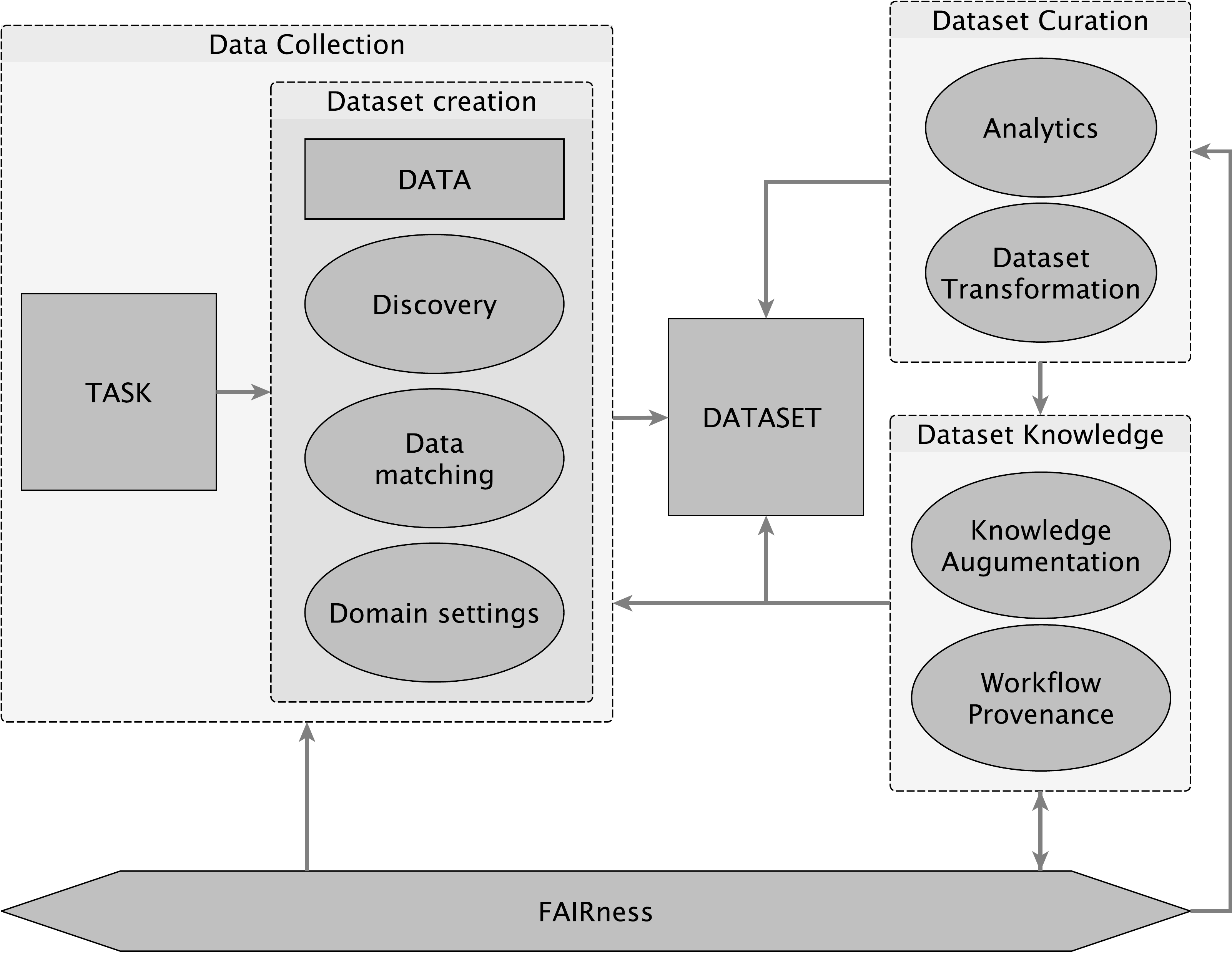}
    \caption{
        The relationship between the main components of our proposal to dataset engineering, the rectangles are instances, and ellipses represent sets of processes.
        FAIRness is a meta element that touches all components.
    }
    \label{fig:datasetengineering}
\end{figure}
We are proposing an approach to the problem of managing data in the discovery process, where the task must guide the data lifecycle. 
Then, from now on, a \textit{dataset} is defined as a set of data structured to perform a well-defined task.
Which can be viewed as a blueprint of the actual data, that can be stored as tables at any time. 
In this text, we do not split the concept of dataset blueprint and dataset tables, but for critical discussion about how to implement such concepts it is paramount.
In Figure~\ref{fig:datasetengineering}, we illustrate the main components of our approach and their relationship. 
Our proposed approach to dataset engineering splits it into three critical dimensions Data Collection, Dataset Curation, and Dataset knowledge.
The data collection is responsible for collecting and organizing the data and creating the datasets from a task. 
It combines many challenges like data discovery and matching with domain settings, which is the set of processes responsible for connecting the data with the specificities of the task for scientific domains, such as material discovery.
The dataset curation component is the set of processes to analyze datasets and enable their evolution.
It focuses on the synergy between analytics and dataset transformation, supporting knowledge augmentation.
The dataset knowledge piece is responsible for extending and connecting a dataset with a knowledge base, managing its lineage, and versioning.
It is also vital feedback on the data collection dimension with the knowledge gathered in all components to improve the entire workflow closing the cycle.
Furthermore, another critical element that we consider is the FAIR principles~\cite{wilkinson:2016:fair-guinding-principles}, which touch all components helping to guide the dataset development.
Although it is a meta part of our framework, in this work, we will discuss FAIRness in the context of the dataset knowledge.

The following sections present each main component from the material discovery perspective. 
This use case is rich in many aspects.
For instance: 
\begin{enumerate*}[label=(\roman*)]
	\item Computational chemistry is a vibrant field with critical issues that create significant challenges for dataset engineering, like how to represent a molecule~\cite{oboyle2012}.
	\item The amount of consumed and created data is enourmos~\cite{ruddigkeit2012}.
	\item The great potential to apply AI in many problems creates a new set of questions directly related to datasets~\cite{suh2020}.
\end{enumerate*}

\section{Data Collection}\label{sec:Data-Collection}

Given that several organizations worldwide continuously produce new data as part
of their discovery and engineering processes, collecting and aggregating such data
comes with several challenges. Here we highlight some challenges that the scientific
community has been trying to address.

\subsubsection{Discovery}
Finding new repositories that host relevant data involves crawling
the Internet. Such a task demands the use of an infrastructure with an efficient storage
stack, a fast network bandwidth, and enough processing power to parse the retrieved pages.
Once potential datasets have been found, one needs to identify the terms of use of such
data and somehow assess the quality of that data. An initiative from bioschemas.org
attempts to improve the website indexing process by defining a markup vocabulary for
websites that host datasets from life sciences~\cite{brickley2019googledatasetsearch}.
On the other hand, many data providers keep falling back to manual curation steps
when processing pipelines flag chemical structures with serious errors. Automatically
assessing the quality of new data remains an open problem that we believe to be of
critical importance for a high-quality data collection task.


\subsubsection{Data matching}
Commonly, researchers attempt to improve upon results published by other organizations.
This means that, once results of that new research are made public, the original data can be augmented by aggregating the new information. Unfortunately, data matching is not a straightforward task, as the following issues observed in material discovery indicate:

\begin{itemize}
    \item Different notations to enumerate chemical structures: mapping between popular text-based notations such as \emph{SMILES}, \emph{SMARTS}, \emph{InChI} and \emph{InChIKey} is required, as publications are free to choose which notation to use
    ~\cite{saldivar2020enumeratingmolecules};
    
    \item Multiple representation for the same molecular graph exist: \emph{SMILES} strings
    are pervasive across data sources, yet one cannot simply resort to string comparison to
    tell if two \emph{SMILES} represent the same molecule. A set of standardization rules allow the
    generation of canonical \emph{SMILES} that converge to the same string. In practice,
    though, standardization rules differs between programs and, consequently, across data
    providers~\cite{bento2020chemblrdkit};
    
    \item Compounds may have different names and several synonyms: for instance, while a publication
    may refer to \textbf{[CH3][CH2][OH]} as \emph{ethanol}, others may refer to that same compound 
    as \emph{alcohol}, \emph{ethyl hydrate}, \emph{anhydrol}, among several different alternative
    names (and possibly written in other languages);
    
    \item Typos and conversion errors: supplemental material, often shared via spreadsheets,
    are prone to conversion errors (such as gene names mistakenly converted to dates by Microsoft Excel~\cite{abeysooriya2021plos}) and typographical errors that further difficult the task
    of automated data matching;
    
    \item Inaccurate cross-references among databases~\cite{dashti2019pubchemeval}.
\end{itemize}

Effectively, the data matching process becomes a pipeline where different techniques apply.
\citet{ongari2022:mofdatamatching}, for instance, attempt matches by conventional name, by
the publication venue and ID, by comparing the molecular graph's substructures, and even
the pore volume of the crystal structures. It is clear that this is an open problem with
several opportunities for improvement.

\subsubsection{Domain settings}
To be useful, data retrieved from any given source repository must be transformed to meet
the needs of the target domain. For instance, transformations like the normalization of
relation units and standardization of compounds depend not only on predefined rules and
conventions but also on ontologies that help establish a mapping from the source data
to the desired output format. Automatically determining which functions provide this mapping
is a key feature of a dataset engineering platform.

\section{Dataset Curation}

The dataset curation step is essential for the understanding of the data as well as the knowledge extraction from it. This step is composed by two different groups which are further detailed: Dataset Analytic, and Dataset Transformation.

\subsection{Dataset Analytic}

The dataset analytic process is composed by different methods which aims to uncover useful insights to experts. This step comprises the following analysis methods: Clustering Analysis, Covering Analysis, Causality Analysis, Data Visualization, Similarity Analysis, and Uncertainty quantification. 

\subsubsection{Clustering Analysis}

The clustering analysis is part of an unsupervised strategy used to discover existing patterns in a given dataset and group objects with similar characteristics given a context. According to \cite{hadipour2022deep}, compounds clustering is vital to validate the diversity of the dataset, identify the similarity and heterogeneity among the objects contained in the dataset, and improve the challenging and costly process of establishing datasets for machine learning tasks \cite{elshawi2018big}. Understanding the categories of the compounds that need to be included in the dataset can significantly reduce the number of molecules that should be screened while, at the same time, ensuring the quality of the dataset. Different clustering techniques can be used at this step, including: CheS-Mapper \cite{gutlein2014ches}, K-Means \cite{nugent2010overview}, Graph-based clustering \cite{tanemura2021autograph}, autocoencoders \cite{hadipour2022deep}, and others. 

\subsubsection{Covering Analysis}

This step regards to the use of evaluation metrics and further insights to understand better the under- or over-generation of the data \cite{tadesse2022mpego}. It also helps to understand their characterizations at different levels of evaluation. Moreover, at this step the completeness of the data can be verified if necessary.  

Therefore, such insights can benefit the quality of the data through improved interactions between machine learning researchers and domain experts in new molecules discovery \cite{tadesse2022mpego}.

\subsubsection{Causality Analysis}

Understanding complicated interactions of chemical components is essential to new molecules discovery \cite{dang2015reactionflow}. Therefore, to detect the causal relations in the molecular structures play essential role for the description of molecular mechanisms and comprehend their functioning \cite{kelly2022review}.

The causality analysis favors the explainability of the dataset and may benefit the process of science discovering through machine learning models \cite{holzinger2021towards}. 

\subsubsection{Data Visualization}

Data visualization is crucial for the dataset analysis and the advance of scientific researches \cite{fox2011changing}. In terms of new molecules discovery field, data visualization enables decision-makers to discover design patterns, comprehend information, and form an opinion about potential new scientific discovery candidates\cite{ekins2016machine}.  

Designing new and better compounds requires understanding of the mechanism by which the molecules exert their biological effects. This also involves consideration of the uncertainty contained in the data, which data visualization helps to provide interpretability of it and allow researches to understand better the nature of  the data \cite{rheingans1999visualization}.

\subsubsection{Similarity Analysis}

Similarity-based methods are part of a feedforward reasoning process that relies on the proximity (in the feature space) of a target object to a given object \cite{angelov2020towards}. In terms of molecules analysis, molecular similarity implies that molecules of ``similar'' structure tend to have similar properties to their analogues \cite{samanta2020vae}. Therefore, a common question that a researcher can make is: ``Given a target molecule $M$ which has a specific chemical activity, can I find the 30 molecules that are most similar to $M$ so I can assess their behavior in a relevant quantitative-structure-activity (QSAR) analysis?''.

The traditional strategy to calculate molecular similarity regards to encode the molecule as a vector of numbers. If fingerprints, for example Extended Connectivity FingerPrinting (ECFP) \cite{rogers2010extended}, are used to produce a vector of bits which that describes molecule structure the Tanimoto similarity is commonly applied. However, as fingerprints just produce binary information regarding to the molecule structure,  molecular descriptors as Mordred descriptor \cite{moriwaki2018mordred} or PaDEL-descriptor \cite{yap2011padel} richer information for the calculation of an improved similarity ranking. These interpretable features provided by molecular descriptors also favors human-in-the-loop as experts can incorporate their knowledge in the data to obtain a set of similar molecules that best fits the context that they are working. 

Ranking metrics also serve an important purpose in evaluating the similarity of compounds. Distinct chemical domains are typically best described by different molecular features. Therefore, to provide experts with guided decision-making capabilities, various techniques can be used to evaluate the quality of any given (encoding, similarity function) pair for some class of compounds \cite{wassenaar2022zzs}.

\subsubsection{Uncertainty Quantification}

Uncertainty quantification is a key component to provide measures of confidence necessary to complex decisions \cite{begoli2019need}. Uncertainty estimation in predictions can have the potential to save considerable time and effort during the decision-making process. Additionally, applications and advances in the field of new molecules discovery has  additional safety requirements that are demanding from scientists \cite{hirschfeld2020uncertainty}.

Indeed, model interpretability including associated confidence in output predictions is recognised as a principal shortcoming of current approaches for new molecules discovery \cite{wan2021uncertainty}. Therefore, better communication of uncertainty positively contributes to the adoption of machine learning systems to accelerate scientific discoveries. 

\subsection{Dataset Transformation}

Data transformation is a key step which embraces the processes of changing the format, structure, or values of data through Interactive AI and Batch Operations. 

\subsubsection{Interactive AI}

This step encompass solutions and algorithms where experts influences AI systems and vice-versa \cite{ccelikok2019interactive}. This includes decision support solutions, recommender systems, and dialogue systems with focus on explainability, interpretability, and interaction leveraging the user experience with the dataset \cite{bellamy2019ai}. This includes the operations of adjudication \cite{schaekermann2020human} and data aggregation \cite{edge2018bringing}. Interactive AI benefits the human-in-the-loop as it allows experts to incorporate useful, meaningful human interaction into the dataset \cite{zanzotto2019human}. This task is specifically important for high stake applications as the dataset engineering to advance science.

\subsubsection{Batch Operations}

Batch operations for data transformation includes processes for data cleaning, reduction, expansion, and generation \cite{fink2009faqs}. Such operations are essential to guarantee the quality of the data and the best usability of them by machine learning approaches and experts. 
Batch operations gives a sense of how data is distributed, both from visual or quantitative perspectives \cite{yu2010exploratory}. Therefore, we can consider that data transformations of variables to ease both interpretation of data analyses and the application statistical and machine learning models to the dataset.

\section{Dataset Knowledge}

Knowledge Augmentation concerns the enhancement of the current knowledge base by acquiring/ingesting new data from other sources, and creating new knowledge by reasoning over the current base and over the new ingested data.
To achieve Knowledge Augmentation, it is required to perform several tasks, like knowledge acquisition, formalization, storage/retrieval, learning, and reasoning~\cite{silva:2022:smart-knowledge-engineering}.

The knowledge base dataset may reside on disparate locations in heterogeneous data stores represented by different data models.
A middleware that provides a seamless interface with an independent data model and data schemes is required to access heterogeneous data stores, like polystore systems~\cite{stonebraker:2015:polystore, ozsu-valduriez:2020:principles-of-data-systems}.

Although exists well-curated, deeply-integrated, special-purpose repositories, many important datasets emerging from traditional, low-throughput bench science do not fit in the data models of these special-purpose repositories.
It results in a diverse, less integrated, data ecosystem, exacerbating the discovery and re-usability of datasets for both humans and computation stakeholders.
As an example, if a researcher wants to compare a dataset resulting from his/her experiment with other datasets, several questions should be answered, such as:
\begin{enumerate*}[label=(\roman*)]
    \item Where might the existing dataset have been published?
    \item How to start the search and using what search tools?
    \item Which characteristics should be used to filter the datasets?
    \item Are the datasets described with metadata, and metadata in what formats?
\end{enumerate*}
After the dataset is found, other questions arise, like:
\begin{enumerate*}[label=(\roman*)]
    \item Can the dataset be downloaded?
    \item What is the data format?
    \item What are the requirements to integrate the data with local data?
    \item Can the data be automatically integrated?
    \item Does the researcher have permission to use the data? Under what license conditions?
\end{enumerate*}
Therefore, it is a grand challenge of data-intensive science to improve knowledge discovery for humans and computational agents in the discovery, access, integration and analysis of task-appropriate scientific data~\cite{wilkinson:2016:fair-guinding-principles}.
In 2016, Wilkinson \etal{}~\cite{wilkinson:2016:fair-guinding-principles} published the FAIR principles, a set of 15 recommendations for improving Findability, Accessibility, Interoperability, and Reusability of digital resources~\cite{jacobsen:2020:fair-principles-considerations}.
The principles are domain-independent and aim to facilitate reuse by humans and machines~\cite{trojahn:2022:fair-core-semantic-metadata-model}.

In another perspective, scientists struggle in performing comprehensive data analyses over their experiments if there is no information collected during the experiment workflow executions.
To overcome this issue, they embrace provenance techniques on their experiments.
Provenance (also referred to as lineage) data management techniques help reproduce, trace, assess, understand, and explain data, models, and their transformation processes~\cite{herschel:2017:suvery-on-provenance,moreau:2008:the-first-profenance-challenge, buneman:2019:data-provenance}.
The provenance research community has evolved significantly to provide for several strategic capabilities, including
    experiment reproducibility~\cite{thavasimani:2016:facilitating-reproducible-research}, 
    user steering (\ie{} runtime monitoring, interactive data analysis, runtime         
        fine-tuning)~\cite{souza:2019:keeping-track-of-user-steering-actions-in-dynamic-workflow}, 
    raw data analysis~\cite{sousa:2016:raw-data-files-through-dataflows}, and
    data integration for multiple workflows generating data in a data lake~\cite{souza:2019:efficient-runtime-capture-of-multiworkflow}.

\section{Final Remarks}
We argue that datasets should have a central role in knowledge-intensive processes, especially in scientific discovery.
Moreover, we define its lifecycle in a task-oriented way, creating a synergy between the natural dataset evolution and its uses.
The tasks and data combination is a powerful driver.
It helps us finding many solutions to reuse data workflows, empower experts in the dataset lifecycle,  and create data tooling.
There are still many questions about the practical issues of datasets. 
However, this dataset definition can significantly impact many real-world problems supporting the acceleration of scientific discovery. 

\bibliography{aaai23}

\end{document}